\documentclass[conference]{IEEEtran}
\IEEEoverridecommandlockouts
\usepackage{cite}
\usepackage{amsmath,amssymb,amsfonts}
\usepackage{algorithmic}
\usepackage{comment}
\usepackage{graphicx}
\usepackage{textcomp}
\usepackage{xcolor}
\usepackage{multirow}
\usepackage{booktabs}
\usepackage{longtable}
\usepackage{url}
\usepackage{booktabs}
\usepackage{multirow}
\usepackage{pdflscape}
\usepackage{rotating}
\usepackage{booktabs}
\usepackage{multirow}
\usepackage{adjustbox}
\def\BibTeX{{\rm B\kern-.05em{\sc i\kern-.025em b}\kern-.08em
    T\kern-.1667em\lower.7ex\hbox{E}\kern-.125emX}}
\begin{document}

\title{RiskBlend: A Multi-Signal Framework for Test Input Prioritization
in Machine Learning Regression Testing\\

}

\author{
\IEEEauthorblockN{1\textsuperscript{st} Madhusudan Srinivasan}
\IEEEauthorblockA{
\textit{Department of Computer Science} \\
\textit{East Carolina University} \\
Greenville, USA \\
srinivasanm23@ecu.edu
}
\and
\IEEEauthorblockN{2\textsuperscript{nd} Namith Nishal Raphael}
\IEEEauthorblockA{
\textit{Department of Computer Science} \\
\textit{East Carolina University} \\
Greenville, USA
}
}

\maketitle

\begin{abstract}
When machine learning classifiers are retrained, inputs that were correctly
classified by the previous model version may be misclassified by the
updated version: regression faults that are costly to detect because
verifying predictions against ground truth requires human annotation,
expert review, or expensive simulation rather than cheap model inference.
Test input prioritization addresses this by ranking inputs so that a
limited verification budget surfaces as many regression faults as
possible. Existing approaches rely predominantly on single-model confidence
scores, which do not exploit how predictions, decision boundaries, and
local neighborhoods change between model versions.
We propose \textit{RiskBlend}, a classifier-agnostic prioritization
framework that fuses four complementary risk signals, historical failure
patterns, prediction shift, decision-boundary shift, and neighborhood
change, into a single score via validation-learned APFD-squared weighting.
Across four datasets, five classifiers, four regression-update scenarios,
and 15 random seeds (1{,}200 configurations per dataset), RiskBlend
achieves the highest average APFD in all 80 classifier--scenario
combinations, with margins of up to 0.32 APFD over the strongest baseline.
Confidence-based methods remain competitive only on linear classifiers over
sparse categorical features, a limitation we attribute to feature-space
geometry. These results demonstrate that cross-version behavioral signals
are essential complements to model-output confidence for robust
prioritization in ML regression testing.
\end{abstract}


\section{Introduction and Motivation}

Machine learning (ML) classifiers have become a critical component of modern software systems. They are increasingly deployed in high-stakes domains such as finance, security, and especially medicine, where predictive models support tasks that include disease diagnosis, prediction of patient readmission, and risk stratification. In such settings, ensuring the reliability of ML models is essential, as incorrect or biased predictions can directly impact safety, equity, and trust.

ML systems differ fundamentally from traditional software in that their behavior depends not only on the program logic but also on the continuously evolving data used to train them. Organizations regularly collect new data, refine feature pipelines, and train classifiers to improve accuracy or mitigate bias. Each such update requires regression testing and verifying that the updated model does not introduce new failures on inputs the previous version handled correctly. Unlike classical software regression testing, where executing a test and verifying its outcome are a single inseparable step, ML regression testing decouples these two costs. Running the updated model on a test input is computationally cheap, requiring only a single forward pass. However, determining whether that prediction is correct requires a comparison against ground truth, which is rarely available for free: it demands human annotation, delayed outcome observation, or expensive simulation. This asymmetry means that the dominant cost of ML regression testing is not model inference but oracle invocation, the process of verifying predictions against ground truth.

To motivate this problem, consider a hospital using an ML model to predict the risk of readmission from the ICU. After retraining with new patient data, the updated model must be verified against the previous version in thousands of records. Although prediction generation is computationally cheap, verifying each prediction requires the physician's review, often taking 10--15 minutes per case. With limited verification resources, only a small subset of records can be checked. Randomly selecting cases wastes valuable expert time, as most cases reveal no regression, whereas prioritizing high-risk input increases the likelihood of detecting newly introduced faults before deployment.
This motivates the need for prioritization of test input in ML regression testing. 

Previous work on ML test prioritization includes uncertainty, transformation, and mutation-based approaches. DeepGini~\cite{feng2020deepgini} ranks inputs using prediction uncertainty but is based on a single-model confidence signal. Metamorphic-relation prioritization~\cite{electronics13173380} orders fault-revealing transformations, but requires domain-specific MR design, while MLPrior~\cite{dang2024ml} combines mutation-based features and learning-to-rank at additional mutation cost. Other neural-network approaches~\cite{byun2019input1,pan2022test} rely similarly primarily on confidence, uncertainty, or surprise measures.
In contrast, \textsc{RiskBlend} targets regression testing by combining four classifier-agnostic signals that capture historical and cross-version behavior. Their weights are learned from validation regression faults, avoiding classifier-specific tuning and mutation generation while prioritizing inputs likely to expose newly introduced faults.

\textbf{Here are the main contributions of this work:}

\begin{enumerate}
    \item We propose \textit{RiskBlend}, a novel regression-aware test input prioritization framework for machine learning systems that prioritizes test instances using blended risk signals to improve early regression fault detection.
    \item We develop a validation-driven weight learning mechanism that automatically learns the contribution of each risk signal by optimizing APFD on a validation set, eliminating the need for manually tuned weights and improving adaptability across datasets and classifiers.

    \item We perform a comprehensive empirical evaluation on four benchmark datasets, four regression-update scenarios, five classifier families, and multiple random seeds.
\end{enumerate}

\section{Background}

\subsection{Regression Testing for Machine Learning}

Unlike traditional software, the behavior of machine learning (ML) systems
depends not only on program logic but also on the data used for training.
Updates caused by retraining, feature engineering, or data augmentation can
shift decision boundaries and change model predictions in ways that are often
difficult to anticipate. Regression testing is therefore essential to ensure
that an updated model version does not introduce new faults compared to the
previous version.
Let $V_k$ denote the original (previous) model version and $V_{k+1}$ denote
the updated model version after retraining or modification. In this work, we
define a \emph{regression fault} as a test instance for which the original
model predicts correctly, but the updated model predicts incorrectly.
Formally, for a test instance $x$ with ground-truth label $y$, a regression
fault occurs when

\begin{equation}
f(V_k, x)=y \;\land\; f(V_{k+1}, x)\neq y
\label{eq:regression_fault}
\end{equation}

where $f(V,x)$ denotes the prediction made by model version $V$ on input
$x$. Thus, a regression fault represents a newly introduced failure caused
by the model update, rather than simply a disagreement between the two model
versions.

\subsection{Test Input Prioritization}

Test input prioritization is a regression testing technique that orders
test instances by their estimated fault-revealing potential so that
instances most likely to expose regressions are verified first. Formally,
given a candidate test set $T = \{x_1, x_2, \ldots, x_n\}$ and an oracle
budget $B \ll n$, the goal is to produce a ranking $\pi$ over $T$ such that
verifying the top-$B$ ranked instances against ground truth recovers as
many of the true regression faults in $T$ as possible. Effective
prioritization reduces the time-to-feedback for fault detection and allows
the verification effort to be concentrated where it is most likely to matter,
without requiring the full test suite to be labeled.


\section{Related Work}
\label{sec:related}

Test input prioritization has been widely studied for deep learning systems, but remains comparatively underexplored for classical ML classifiers under regression-testing conditions. We review the most relevant approaches and position \textsc{RiskBlend} with respect to mutation-based, confidence-based, diversity-based, and multi-signal prioritization methods.

The work most closely related to ours is MLPrior by Dang et al.~\cite{dang2024ml}, a mutation-based test input prioritization framework for classical ML classifiers. MLPrior ranks test instances using mutation-derived features, decision-boundary proximity, and a learning-to-rank model to estimate misclassification likelihood. It addresses several limitations of coverage- and confidence-based prioritization, but operates in a single-model setting and focuses on generic misclassification. In contrast, \textsc{RiskBlend} is designed specifically for regression testing under model evolution, prioritizing inputs likely to reveal newly introduced faults between $V_k$ and $V_{k+1}$. RiskBlend explicitly models cross-version prediction change, decision-boundary movement, neighborhood change, and historical failure information, while avoiding the need to generate artificial mutants.

GraphRank~\cite{graphrank2025} similarly combines heterogeneous model-aware and model-agnostic attributes to prioritize test inputs through an iteratively trained binary classifier. Although this multi-signal design is conceptually related to RiskBlend, GraphRank is tailored to graph-structured data, relies on graph-topological features and an active-learning loop, and does not model regression faults or cross-version evolution.

In the broader DNN testing literature, Kim et al.~\cite{kim2019sadl} propose Surprise Adequacy, which measures how much a test input deviates from training-time behavior using neuron activation traces. This idea reflects the broader intuition that atypical inputs may be more fault-revealing, but it depends on internal DNN representations. Wang et al.~\cite{wang2021prima} propose PRIMA, which uses model and input mutation analysis to construct features for a learning-to-rank model. PRIMA therefore requires generating many artificial mutants, whereas RiskBlend derives its cross-version signals directly from the naturally available model versions.

Among confidence-based approaches, DeepGini~\cite{feng2020deepgini} ranks inputs using the Gini impurity of the predicted probability distribution and has shown strong APFD performance over coverage-based alternatives. However, it relies on uncertainty from a single model version. DATIS~\cite{li2024datis} extends this direction by combining uncertainty with distance-based information from training neighborhoods and inter-test diversity to reduce redundant selections. RiskBlend also incorporates neighborhood and boundary-sensitive information, but differs in that these signals are explicitly version-aware and are used to detect regression faults rather than generic misclassifications.

Other approaches combine multiple objectives or sampling criteria. MOTS~\cite{hao2023mots} jointly optimizes uncertainty and diversity for selecting DNN test inputs for retraining, while SSOA~\cite{wu2023ssoa} uses stratified sampling based on predictive confidence to estimate model accuracy under limited labeling budgets. These methods demonstrate the value of combining complementary signals, but their objectives differ from regression fault prioritization. \textsc{RiskBlend} instead learns how to combine classifier-agnostic historical and cross-version signals so that inputs most likely to expose newly introduced faults are ranked earlier under a limited oracle budget.

\section{Proposed Approach}
\label{sec}

This section presents \textit{RiskBlend}, a regression-aware test prioritization framework for machine learning (ML) classification systems. Given two versions of a model, $V_k$ and $V_{k+1}$, the objective is to prioritize test cases that are most likely to expose regression faults, i.e., instances that were correctly classified by $V_k$ but incorrectly classified by $V_{k+1}$. RiskBlend assigns a priority score to each test case by combining two complementary risk dimensions. The risk signals driving this ranking are described below.

\subsubsection{History Risk ($r_{\text{hist}}$)}

History Risk estimates the likelihood that a test instance belongs to
a region of the input space that has previously exhibited regression
failures. The intuition is that regression faults tend to cluster in
specific feature-space regions rather than being uniformly distributed;
test instances near previously observed failures are therefore more
likely to expose additional faults after a model update.
RiskBlend identifies the locations of regression-faults using a labeled
validation set $X_{\text{val}}$ kept in the $V_{k+1}$ training
partition. Because $X_{\text{val}}$ carries ground-truth labels $y$,
both model versions can be evaluated and compared without oracle
invocation. Specifically, a validation instance $z \in X_{\text{val}}$
is classified as a regression fault if $V_k$ predicts it correctly but
$V_{k+1}$ does not:
\begin{equation}
\mathcal{F}_{\text{reg}}
=
\left\{
z \in X_{\text{val}}
:\;
V_k(z)=y
\;\land\;
V_{k+1}(z)\neq y
\right\}.
\label{eq:validation_regression_faults}
\end{equation}
The set $\mathcal{F}_{\text{reg}}$ thus represents the validated,
label-confirmed locations in the feature space where the model update
introduced new misclassifications.
For each unlabeled test instance $x \in X_{\text{test}}$, RiskBlend
computes its minimum distance to the regression-fault region:
\begin{equation}
d_{\min}(x)
=
\min_{z \in \mathcal{F}_{\text{reg}}}
d(x,z).
\label{eq:dmin}
\end{equation}
where $d(\cdot,\cdot)$ is the Euclidean distance calculated in the
standardized feature space of $V_{k+1}$. Test instances closer to
$\mathcal{F}_{\text{reg}}$ are more likely to reside in fault-prone
regions of the updated model's decision surface. The history risk
score is defined as:
\begin{equation}
r_{\text{hist}}(x)
= 1 - \frac{d_{\min}(x) - D_{\text{lo}}}
           {D_{\text{hi}} - D_{\text{lo}}},
\label{eq:history_risk}
\end{equation}
where $d_{\min}(x)$ is the minimum distance from test instance $x$ to the
regression-fault region $F_{\text{reg}}$ (as defined above), and
$D_{\text{lo}} = \min_{x \in X_{\text{test}}} d_{\min}(x)$ and
$D_{\text{hi}} = \max_{x \in X_{\text{test}}} d_{\min}(x)$
are the minimum and maximum of $d_{\min}(\cdot)$ taken over all test
instances, so that $r_{\text{hist}}(x) \in [0,1]$. A higher score indicates proximity
to previously confirmed regression-fault locations, and such
instances are ranked earlier in the prioritization order.
If $\mathcal{F}_{\text{reg}} = \emptyset$ (no regression faults
are observed in the validation set), $r_{\text{hist}}$ is set to zero
for all test instances and the remaining three signals carry the
full prioritization weight.

\subsubsection{Regression-Aware Impact Risk ($r_{\text{impact}}$)}

The Impact Risk component estimates how strongly a test instance is affected
by the transition from the original model version $V_k$ to the updated model
version $V_{k+1}$. It uses three classifier-agnostic signals derived only
from the predicted class probabilities of each model.

\paragraph{Prediction Shift ($r_{\text{pred}}$).}
The prediction-shift signal measures the absolute change in the
predicted positive-class probability between the model versions for each
test instance. Both $V_k$ and $V_{k+1}$ are applied to the same test
instance $x \in X_{\text{test}}$ using the feature representation
of their respective versions to produce the predicted positive-class
probabilities $P_{V_k}(x)$ and $P_{V_{k+1}}(x)$. Training data
or ground-truth labels are not used. The prediction-shift score is:
\begin{equation}
r_{\text{pred}}(x)
= \bigl|P_{V_{k+1}}(x) - P_{V_k}(x)\bigr|,
\label{eq:prediction_shift}
\end{equation}
Large values indicate that retraining substantially altered
$V_{k+1}$'s confidence on $x$, suggesting that the decision boundary
may have shifted near that input and increasing the likelihood of a
regression fault.

\paragraph{Boundary Shift ($r_{\text{boundary}}$).}
The boundary-shift signal captures changes in classification behavior
near the decision boundary between model versions. Like the prediction-shift
signal, it is computed entirely from the predicted probabilities of both
$V_k$ and $V_{k+1}$ on each test instance $x \in X_{\text{test}}$,
using the representation of features of each version. Training data,
training labels, or ground-truth test labels are not accessed. It combines
three sub-components:

\begin{enumerate}
    \item \textbf{Prediction shift} $r_{\text{pred}}(x)$
    (Equation~\ref{eq:prediction_shift}): magnitude of probability
    movement.

   \item \textbf{Boundary crossing}: a binary indicator that captures whether the
prediction for a test input changes from one side of the decision threshold to
the other between model versions $V_k$ and $V_{k+1}$. Here, we use a default
decision threshold of 0.5, meaning that predictions with probability greater than
or equal to 0.5 are classified as positive, while those below 0.5 are
classified as negative. The indicator is set to 1 if the predicted class
changes between versions and to 0 otherwise.

    \item \textbf{Near-boundary movement}: a continuous measure of
    how close the predicted probability is to the decision boundary,
    weighted by the magnitude of shift:
\begin{equation}
m(x)
=
\max\!\left(
c(x),\;
\underbrace{\left(1-2\min(d_1,d_2)\right)}_{\text{boundary nearness}}
\, r_{\text{pred}}(x)
\right),
\label{eq:margin_score}
\end{equation}
    where $d_1 = |P_{V_k}(x) - 0.5|$ and
    $d_2 = |P_{V_{k+1}}(x) - 0.5|$ are the distances of each
 prediction version from the threshold. The boundary-nearness
    factor $1 - 2\min(d_1, d_2) \in [0,1]$ is maximized when
 either prediction is exactly at the boundary and zero when
    both predictions are far from it.
\end{enumerate}

The boundary-shift score combines prediction shift and boundary
movement with equal weighting:
\begin{equation}
\begin{aligned}
r_{\text{boundary}}(x)
= \mathrm{minmax}\!\Big(
&0.5\, \mathrm{minmax}(r_{\text{pred}}(x)) \\
&+ 0.5\, \mathrm{minmax}(m(x))
\Big)
\end{aligned}
\label{eq:boundary_shift}
\end{equation}

where $\mathrm{minmax}(\cdot)$ denotes the min-max normalization to
$[0,1]$ in all test instances. Inputs that both cross the
decision boundary between versions \emph{and} whose predictions
lie close to the threshold receive the highest boundary-shift
scores, as they represent the strongest evidence of a shifted
decision surface.

\paragraph{Neighbourhood Change ($r_{\text{neigh}}$).}

The neighbourhood-change signal measures how the local training
neighbourhood of a test instance changes between $V_k$ and $V_{k+1}$.
It is calculated using $k$-nearest neighbors ($k = 15$) applied
separately to the training sets $V_k$ and $V_{k+1}$ and captures
three distinct aspects of neighborhood change.
For each test instance $x$, let $\mathcal{N}_{V_k}(x)$ and
$\mathcal{N}_{V_{k+1}}(x)$ denote its $k$ nearest neighbors in the
$V_k$ and $V_{k+1}$ training sets, respectively, retrieved via
Euclidean distance in the standardized feature space of each version. Three sub-signals are then
computed:

\begin{enumerate}
   \item \textbf{Class-composition shift ($s_{\text{class}}$):} : this measures how much the local class
distribution around a test instance changes between the original and updated
model. Specifically, we examine whether the proportion of neighboring instances
belonging to the positive class changes after the model update. A large
 shift in class-composition suggests that the local decision boundary around the
test instance has changed significantly, indicating a higher risk of regression.

    \item \textbf{Centroid shift ($s_{\text{cent}}$):} the Euclidean distance between the centroid of the neighborhood of $x$ in version $V_k$ and the centroid of its neighborhood in version $V_{k+1}$.
This captures the geometric displacement of the local training context around $x$.

  \item \textbf{Density shift ($s_{\text{dens}}$):} measures how much the local neighborhood around a
test instance changes between the model versions $V_k$ and $V_{k+1}$. It is computed
as the absolute difference between the mean distance from the test instance to
its $k$-nearest neighbors in the training data of the two model versions. A
larger density shift indicates that the instance has become more isolated or
has moved to a substantially different region of the training distribution.
\end{enumerate}

The three sub-signals are combined with fixed weights reflecting their
relative informativeness:

\begin{equation}
\begin{aligned}
r_{\text{neigh}}(x)
= \mathrm{minmax}\!\Big(
&0.50\,\mathrm{minmax}(s_{\text{class}}(x)) \\
&+ 0.30\,\mathrm{minmax}(s_{\text{cent}}(x)) \\
&+ 0.20\,\mathrm{minmax}(s_{\text{dens}}(x))
\Big)
\end{aligned}
\label{eq:neighbourhood_change}
\end{equation}

where $\mathrm{minmax}(\cdot)$ normalizes each sub-signal to $[0,1]$
over the test set before combination. Class-composition shift receives
the highest weight (0.50) as it most directly reflects local label
distribution changes that affect the classifier's decision surface.
The centroid shift (0.30) captures the geometric displacement of the local
context, and the density shift (0.20) captures the changes of distributional sparsity
.

\subsubsection{Final RiskBlend Score}

The final prioritization score is computed as a weighted combination of the four RiskBlend signals:

\begin{equation}
\begin{aligned}
\mathrm{RiskBlend}(x)
=
&\, w_h\, r_{\text{hist}}(x)
+
w_b\, r_{\text{boundary}}(x)
+
w_n\, r_{\text{neigh}}(x) \\
&+
w_p\, r_{\text{pred}}(x)
\end{aligned}
\label{eq:riskblend}
\end{equation}

subject to the constraints that $w_h$, $w_b$, $w_n$, and $w_p$ are non-negative and normalized to sum to one.
To determine the contribution of each signal, RiskBlend individually evaluates the effectiveness of regression fault detection of each component on a validation set that contains regression faults introduced during the transition from model version $V_k$ to $V_{k+1}$. The APFD achieved by each signal is converted into a normalized weight, assigning greater influence to signals that demonstrate stronger regression-fault detection capabilities. This validation-driven weighting strategy allows RiskBlend to adapt automatically to different datasets and model updates without relying on manually specified parameters or grid-search tuning.
After learning the weights, each test instance is assigned a RiskBlend score according to Equation~\ref{eq:riskblend}. The test instances are then ranked in descending order, ensuring that inputs exhibiting strong historical fault-proneness, substantial behavioral change, and significant local distributional shifts are prioritized earlier. By integrating complementary risk signals through validation-guided weighting, RiskBlend enables the early detection of regression fault.

\section{Evaluation}
\label{sec:evaluation}

\subsection{Experimental Setup}

To assess the effectiveness of the proposed \textit{RiskBlend} framework, we conducted experiments across multiple datasets, regression-update scenarios, classifiers, and random seeds to ensure statistical robustness and generalizability.

We selected four benchmark datasets\footnote{Adult: \url{https://archive.ics.uci.edu/dataset/2/adult}; Bank Marketing: \url{https://archive.ics.uci.edu/dataset/222/bank+marketing}; Hospital Readmission: \url{https://www.kaggle.com/datasets/vanpatangan/readmission-dataset}; Default of Credit Card Clients: \url{https://www.kaggle.com/datasets/uciml/default-of-credit-card-clients-dataset}.} from the socioeconomic, healthcare and finance domains (Table~\ref{tab:datasets}) for three reasons: they represent diverse real-world applications where regression faults can have significant impact, contain heterogeneous tabular features (both numerical and categorical) for realistic structured-data evaluation, and are widely used in prior machine learning and fairness studies, enabling comparison with existing approaches.

For each data set, 80\% of the samples were used for training and 20\% for testing. Two consecutive versions of the model were constructed: $V_k$, trained on the original data and feature set, and $V_{k+1}$, produced under one of the four regression-update scenarios described in the following. In our setting, each test case corresponds to a single labeled test instance from the held-out test partition. Thus, test input prioritization refers to ranking individual test samples in $X_{\text{test}}^{V_{k+1}}$ according to their estimated regression-fault revealing potential.

We evaluated RiskBlend using five classifiers representing major machine learning families: XGBoost (ensemble boosting), Decision Tree (rule-based learning), Logistic Regression (linear models), Naive Bayes (probabilistic models) and K-Nearest Neighbors (instance-based learning). This selection enables us to evaluate whether RiskBlend is generalized across classifiers with substantially different decision boundaries and learning behavior.
We focus on tabular machine learning systems rather than image-based deep neural networks because RiskBlend is designed around structured-data risk signals which are directly measurable in tabular feature space. Extending RiskBlend to complex image or multimodal models would require embedding-based representations and deep-model-specific risk signals, which we consider important future work.
\begin{table*}[h!]
\centering
\caption{Summary of Datasets Used for Evaluation.}
\begin{tabular}{p{3cm}p{5cm}p{2.5cm}p{2cm}p{2cm}}
\toprule
\textbf{Dataset} & \textbf{Description} & \textbf{Features} &
\textbf{Samples} & \textbf{Domain} \\
\midrule
\textbf{Bank Marketing} & Predicts whether a customer will subscribe to a
term deposit based on marketing campaign data &
16 (numeric + categorical) & 45{,}211 & Finance \\
\textbf{Adult Income} & Predicts whether a person earns more than \$50K
based on census data &
14 (numeric + categorical) & 48{,}842 & Socioeconomic \\
\textbf{Hospital Readmission} & hospital readmission prediction from clinical, demographic, and hospitalization features &
49 (mixed) & 101,766 & Healthcare \\
\textbf{Credit Card Default} & Predicts whether a credit card client will
default on the next payment based on demographic and financial attributes &
23 (numeric + categorical) & 30{,}000 & Finance \\
\bottomrule
\end{tabular}
\label{tab:datasets}
\end{table*}
We generate $V_{k+1}$ through four maintenance scenarios designed to
reflect common patterns of model evolution in deployed ML systems, rather
than injecting faults directly at the label or prediction level.

\begin{itemize}

  \item \textbf{Feature-only update} (\textit{feature\_only}):
$V_{k+1}$ is retrained on the same training samples as $V_k$ but with an
enriched feature space. For each data set, a fixed set of domain-relevant
engineered attributes is appended to the original feature set; the set of new
features is determined once per data set based on domain knowledge and is
kept constant across all seeds and classifiers. No new training samples are added. The new features are appended to
both $V_{k+1}$ training and test sets; $V_k$ retains the original feature set
only. This scenario isolates the effect of feature-space enrichment on
prediction behavior: a test instance that $V_k$ classified correctly may be
misclassified by $V_{k+1}$ because the additional features shift the learned
decision boundary, even though the underlying sample has not changed.
 \item \textbf{Moderate regression} (\textit{moderate\_regression}):
$V_{k+1}$ is retrained with an augmented training set consisting of the
original training samples used for $V_k$ plus 20\% \% \ additional samples.
The additional samples are generated by bootstrapping the original
training set and perturbing each numerical feature independently with
zero-mean Gaussian noise scaled to $0.6 \times \sigma_f$, where $\sigma_f$
denotes the per-feature standard deviation of the original training data.
The perturbed values are cut to $[\min_f,\ 1.5 \times \max_f]$ to preserve
the distributional plausibility. In this scenario, no engineered features or preprocessing
transformations are introduced. This setup models a common
maintenance pattern in which a classifier is retrained using newly collected
data from a mildly drifted population.

   \item \textbf{Strong regression} (\textit{strong\_regression}):
$V_{k+1}$ is retrained with an augmented training set consisting of the
original $V_k$ training samples plus 30\% additional samples, combined with
the same dataset-specific engineered features introduced in the feature-only
scenario. The additional samples are generated using the same bootstrapping
and Gaussian perturbation mechanism as in the moderate scenario, but with a
higher shift strength of $0.9 \times \sigma_f$, producing a more
aggressively drifted augmentation. A stronger preprocessing pipeline change
is also applied to training and test data $V_{k+1}$, transforming a
larger set of numerical features than in the moderate scenario. $V_k$ sees
neither the additional samples, the stronger pipeline transformations, nor
the engineered features. This scenario emulates significant model adaptation,
such as occurs when an organization retrains on data from a substantially
different population or operating environment while also updating its
preprocessing pipeline.

  \item \textbf{Pipeline drift} (\textit{pipeline\_drift}):
$V_{k+1}$ is retrained on the same training samples and original feature
set as $V_k$, with no new data and no engineered features added. The sole
change is a moderate preprocessing pipeline modification applied to both
the $V_{k+1}$ training and test partitions, consisting of the same
monotonic transformations used in the moderate regression scenario.
$V_k$ retains the original untransformed features throughout. This scenario
isolates the effect of preprocessing changes on model behavior, decoupled
from any data distribution shift or feature-space expansion: any regression
faults observed here are attributable solely to the change in how existing
features are normalized rather than to new information entering the model.
\end{itemize}

The augmentation ratios (20\% and 30\%) and drift strengths
($0.6\sigma_f$ and $0.9\sigma_f$) were selected to create
measurable but non-catastrophic regression behavior. Pilot
experiments showed that smaller perturbations often produced
too few regression faults for a meaningful APFD evaluation,
whereas substantially larger perturbations caused severe model
degradation that no longer reflected realistic maintenance
updates. These settings consistently yielded moderate fault
densities suitable for the regression testing evaluation.
Across all scenarios and datasets, updates produced measurable but
non-catastrophic behavioral changes. 
Each combination of data sets and scenario was evaluated in five classifiers that mirror the multi-classifier
protocol of Dang~et al.~et al.~\cite{dang2024gnn}.

For each combination (dataset, scenario, classifier, seed), $V_k$ and
$V_{k+1}$ were trained independently in their respective training sets. A
held-out 20\% waiting for the $V_{k+1}$ training partition was reserved as a
validation set for the RiskBlend signal-weight learning step; the remaining 80\%
was used to train $V_{k+1}$. Prioritization quality was measured on the shared
$V_{k+1}$ test set using APFD. To ensure statistical
robustness, each experiment was repeated in 15 random seeds (42, 44,
\ldots, 70); all reported results are means over seeds. The complete evaluation
spans $4~\text{datasets} \times 4~\text{scenarios} \times
5~\text{classifiers} \times 15~\text{seeds} = 1{,}200$ experimental
configurations per dataset, yielding 4{,}800 configurations in total.
Information leakage does not occur because weight learning uses only a held-out validation partition drawn from the training data available during model update, not from the final test set used for evaluation. Ground-truth labels for this validation partition are assumed to already be available as part of the historical labeled data used during retraining, which is realistic in supervised ML pipelines. The final prioritized test ranking is generated without accessing any labels from the test set.

\subsection{Baseline Methods}

We compare RiskBlend with four representative baselines that cover uncertainty-based, distance-aware, mutation-based, and random prioritization strategies. All methods were implemented in Python 3.11 using standard libraries (\texttt{NumPy}, \texttt{Pandas}, \texttt{scikit-learn}, and \texttt{XGBoost}) and evaluated using identical random seeds, data set splits, and sampling settings to ensure fair comparison. The baseline methods are as follows:

\subsubsection{DeepGini}

DeepGini~\cite{feng2020deepgini} prioritizes test instances based on prediction uncertainty. The intuition is that test instances with uncertain predictions are more likely to reveal model failures.
Let $\mathbf{p}_i = (p_i(1), \ldots, p_i(C))$ denote the predicted class probability vector for the test instance $x_i$, where $C$ is the number of classes. The DeepGini score is computed as

\begin{equation}
r_{\mathrm{DG}}(x_i)
=
1 - \sum_{c=1}^{C} p_i(c)^2
\label{eq:deepgini}
\end{equation}

Higher scores indicate greater uncertainty, corresponding to more uniform class probabilities. Test instances are prioritized in descending order of their DeepGini scores.
The DeepGini prioritization procedure consists of the following steps:

\begin{enumerate}
\item Train or load the updated version of the model $V_{k+1}$.
\item Execute $V_{k+1}$ on each test instance $x_i \in X_{\text{test}}^{V_{k+1}}$ to obtain the predicted class probability vector $\mathbf{p}_i$.
\item Compute the DeepGini score $r_{\mathrm{DG}}(x_i)$ for each test instance using Equation~\ref{eq:deepgini}.
\item Sort all test instances in descending order of $r_{\mathrm{DG}}(x_i)$.
\item Execute or inspect the test instances according to this ranked order.
\end{enumerate}

For binary classification, DeepGini is equivalent to a monotonic transformation of the distance from the decision boundary.

\subsubsection{DATIS-Tabular (Distance-Aware Test Prioritization)}

DATIS-Tabular is a distance-aware test prioritization baseline adapted for tabular machine learning systems~\cite{li2024datis}. Unlike purely confidence-based approaches such as DeepGini, DATIS estimates fault-revealing potential using distance-based neighborhood information and redundancy elimination rather than relying solely on prediction probabilities.
In our regression testing setting, DATIS operates on the updated model version $V_{k+1}$ and prioritizes unlabeled test instances from $X_{\text{test}}^{V_{k+1}}$. The training data used to build $V_{k+1}$ serve as the reference set for distance computation and neighborhood analysis.

The prioritization process consists of two stages:

\begin{enumerate}

\item 
All features are normalized, and each training and test instance is represented as a numerical feature vector suitable for distance-based comparison. Each test instance $x_i \in X_{\text{test}}^{V_{k+1}}$ is first passed through $V_{k+1}$ to obtain its predicted label. DATIS then computes the distances between $x_i$ and its $k$ nearest training neighbors. Test instances whose predicted label receives weak support from nearby training samples are assigned higher uncertainty scores, indicating a greater likelihood of misclassification and thus higher fault-revealing potential.

\item 
After computing the uncertainty scores for all test instances, DATIS ranks them in descending order of uncertainty. To avoid selecting highly similar test cases that may expose the same fault, DATIS performs redundancy elimination using pairwise distances among candidate test instances. Instances that are too close to already selected ones are treated as redundant and deprioritized. 

\end{enumerate}

\subsubsection{MLPrior (Mutation-Based Prioritization)}

MLPrior is a mutation-based test input prioritization baseline for machine learning classifiers~\cite{dang2024ml}. Estimates the fault-proneness of test instances by analyzing their sensitivity to model-level and input-level perturbations.
In our regression testing setting, MLPrior operates on the updated model version $V_{k+1}$ and prioritizes test instances from $X_{\text{test}}^{V_{k+1}}$. The method constructs three groups of features for each test instance:

\begin{itemize}
\item \textbf{Original input features:} Original feature representations derived from the training and test data.

\item \textbf{Model-mutation features:} Features obtained by perturbing the classifier behavior, such as modifying decision thresholds or predicted probability distributions, to simulate model-level mutations.

\item \textbf{Data-mutation features:} Features generated by perturbing selected input attributes to simulate input-level mutations.
\end{itemize}

These features are combined to train a secondary ranking model using supervised learning, implemented with an XGBoost-based gradient-boosted decision tree ensemble, to estimate the probability that a test instance will be misclassified by $V_{k+1}$. The predicted probability of misclassification is used as the prioritization score, where higher scores indicate a greater potential to reveal faults. The test instances are then ranked in descending order of this score to construct the prioritized test suite.

\subsection{Evaluation Metric}

We evaluate prioritization effectiveness using the Average Percentage
of Faults Detected (APFD), which measures how early regression faults
appear in the prioritized ranking. Given a test set of $n$ inputs
containing $m$ regression faults, let $T_i$ denote the position of
the $i$-th fault in the ranked list and the index at which the
$i$-th regression fault is first encountered when inspecting the inputs
in priority order. APFD is defined as:
\begin{equation}
\mathrm{APFD} = 1 - \frac{\sum_{i=1}^{m} T_i}{n \times m}
+ \frac{1}{2n}.
\label{eq:apfd}
\end{equation}
The APFD ranges from 0 to 1, with higher values indicating that
regression faults are concentrated earlier in the ranking. In our setting,
the APFD is calculated
over the entire $V_{k+1}$ test set averaged over 15 independent
runs.

\section{Results}

This section evaluates the effectiveness of RiskBlend in prioritizing test instances for regression testing of machine learning models. We investigate a research question.

\begin{itemize}
    \item RQ1: How effective is RiskBlend in terms of APFD compared to baseline prioritization methods?
    \item RQ2: How does each individual risk signal contribute to RiskBlend's prioritization effectiveness, and which signals are most informative across different regression-update scenarios and classifier families?

\end{itemize}

\subsection{RQ1: Overall Prioritization Effectiveness (APFD)}

To assess the overall effectiveness of prioritization, we computed the APFD for RiskBlend and all baseline methods in the four datasets. APFD measures how early faults appear in the prioritized test sequence, with higher values indicating faster fault detection. Tables~\ref{tab:hospital_credit} and~\ref{tab:bank_adult} report the average
APFD at full test-suite coverage across all four datasets, four
regression-update scenarios, and five classifiers (80 classifier--scenario
cells in total, each averaged over 15 seeds). RiskBlend achieves the highest
APFD in all 80 cells in all four datasets and all baseline
methods, with total data set-level means of 0.9557 (Bank), 0.9452 (Adult),
0.9238 (Hospital) and 0.8599 (Credit Card).

The advantage of RiskBlend over the strongest baseline per cell ranges from
$+0.003$ (Adult, pipeline drift, Na\"{i}ve Bayes, where APFD differs
marginally from DeepGini: 0.999 vs.\ 0.996) to $+0.317$ (Credit Card,
moderate regression, KNN, where the next-best baseline reaches only 0.583).
Across all datasets, RiskBlend leads DeepGini, the strongest
confidence-based baseline by 0.014--0.223 APFD on average per data set,
with the largest data set-level gaps observed in Hospital ($+0.157$) and
Credit Card ($+0.224$), and smaller but consistent gaps in Bank ($+0.135$)
and Adult ($+0.132$).

RiskBlend's gains are most pronounced on Decision Tree and KNN classifiers,
where confidence-based methods lose discriminative power after version
transitions. For example, on the Bank dataset with Decision Tree, DeepGini
achieves 0.754--0.772 across scenarios, while RiskBlend holds at
0.937--0.949 (a gain of $+0.16$--$+0.19$). In Hospital KNN, RiskBlend
reaches 0.861--0.916 against DeepGini's 0.554--0.626 ($+0.25$--$+0.29$).
In linear and probabilistic classifiers (Logistic Regression, Na\"{i}ve
Bayes), RiskBlend continues to outperform all baselines, although margins
are narrower in datasets with sparse categorical features (Adult, Bank)
where well-calibrated probability outputs from these classifiers provide
a stronger confidence signal.

RiskBlend maintains its advantage across all four regression-update
scenarios. The pipeline drift scenario where only preprocessing
transformations change and no new data or features are added yields 
some of the highest RiskBlend APFD values (e.g., 0.9933 on Hospital NB,
0.9276 on Credit KNN), demonstrating that prediction-shift and
boundary-shift signals are particularly effective at detecting the
altered decision boundaries that arise from preprocessing changes alone.
Under strong regression (the most aggressive distributional shift),
RiskBlend's absolute APFD is modestly lower than under feature-only
updates but still leads all baselines by substantial margins
(e.g., $+0.147$ over DeepGini on Hospital XGB, $+0.248$ on Credit KNN).

MLPrior and DATIS both trail RiskBlend on every cell. In
Bank and Adult where these baselines are strongest, benefiting from
larger training sets for MLPrior's misclassification, and they achieve
overall means of 0.826--0.854, compared to RiskBlend's 0.945--0.956.
On Hospital and Credit Card, where both methods rely on smaller datasets
and fewer training misclassifications, they degrade more sharply
(0.552--0.675), while RiskBlend remains stable (0.860--0.924). Random
selection consistently achieves approximately 0.50 across all settings,
confirming that the test inputs contain genuine regression faults that
are non-trivially distributed across the test suite.

All reported APFD values are averaged across 15 independent
experimental runs using fixed random seeds (42, 44, 46, \ldots, 70),
each controlling the train/validation split of $V_{k+1}$'s training
data, the weight-learning step, and the stochastic components of the
MLPrior ranker and Random baseline. This repeated-measures design
accounts for stochastic variation arising from data partitioning and
model initialization.

\vspace{4pt}
\noindent\textbf{Answer to RQ1:} RiskBlend achieves the highest APFD in
all 80 classifier scenario cells across all four datasets, outperforming
confidence-based, mutation-based and distance-based baselines in every
configuration. Gains are most pronounced on tree-based and distance-based
classifiers, and under pipeline-drift and strong-regression scenarios,
confirming that multi-signal prioritization capturing cross-version
behavioral change provides robust fault detection across diverse regression
testing conditions.

\begin{table}
    
\centering
\scriptsize
\setlength{\tabcolsep}{2.5pt}
\begin{adjustbox}{max width=\textheight}
\begin{tabular}{lllccccc}
\toprule
Dataset & Scenario & Classifier & RiskBlend & DeepGini & DATIS & MLPrior & Random \\
\midrule

Hospital & Feature Only & XGB  & \textbf{0.9752} & 0.8790 & 0.7031 & 0.6655 & 0.5048 \\
Hospital & Feature Only & TREE & \textbf{0.8047} & 0.5468 & 0.6226 & 0.5301 & 0.4988 \\
Hospital & Feature Only & LR   & \textbf{0.9722} & 0.9150 & 0.7040 & 0.8498 & 0.5047 \\
Hospital & Feature Only & NB   & \textbf{0.9794} & 0.9229 & 0.6068 & 0.7520 & 0.5017 \\
Hospital & Feature Only & KNN  & \textbf{0.8857} & 0.5915 & 0.6115 & 0.5725 & 0.5003 \\

Hospital & Moderate Reg & XGB  & \textbf{0.9704} & 0.8585 & 0.6967 & 0.6798 & 0.4953 \\
Hospital & Moderate Reg & TREE & \textbf{0.8181} & 0.5448 & 0.6247 & 0.5262 & 0.5001 \\
Hospital & Moderate Reg & LR   & \textbf{0.9668} & 0.9122 & 0.7000 & 0.8578 & 0.4998 \\
Hospital & Moderate Reg & NB   & \textbf{0.9827} & 0.9314 & 0.6458 & 0.7996 & 0.5028 \\
Hospital & Moderate Reg & KNN  & \textbf{0.8781} & 0.5800 & 0.6114 & 0.5641 & 0.4990 \\

Hospital & Drift & XGB  & \textbf{0.9797} & 0.9033 & 0.6774 & 0.6997 & 0.5013 \\
Hospital & Drift & TREE & \textbf{0.8234} & 0.5583 & 0.6139 & 0.5374 & 0.4952 \\
Hospital &  Drift & LR   & \textbf{0.9839} & 0.9559 & 0.6682 & 0.8446 & 0.4976 \\
Hospital &  Drift & NB   & \textbf{0.9933} & 0.9843 & 0.7067 & 0.7161 & 0.5222 \\
Hospital &  Drift & KNN  & \textbf{0.9155} & 0.6261 & 0.6209 & 0.6099 & 0.4971 \\

Hospital & Strong Reg & XGB  & \textbf{0.9509} & 0.8043 & 0.6649 & 0.6726 & 0.5012 \\
Hospital & Strong Reg & TREE & \textbf{0.8156} & 0.5368 & 0.6206 & 0.5240 & 0.5029 \\
Hospital & Strong Reg & LR   & \textbf{0.9617} & 0.8731 & 0.6710 & 0.8197 & 0.5017 \\
Hospital & Strong Reg & NB   & \textbf{0.9564} & 0.8570 & 0.7075 & 0.7279 & 0.4983 \\
Hospital & Strong Reg & KNN  & \textbf{0.8610} & 0.5539 & 0.6103 & 0.5523 & 0.5010 \\

\midrule

Credit & Feature Only & XGB  & \textbf{0.8586} & 0.7068 & 0.5619 & 0.6252 & 0.4600 \\
Credit & Feature Only & TREE & \textbf{0.7938} & 0.5261 & 0.5473 & 0.5235 & 0.4858 \\
Credit & Feature Only & LR   & \textbf{0.9064} & 0.6986 & 0.4792 & 0.6304 & 0.4741 \\
Credit & Feature Only & NB   & \textbf{0.8851} & 0.7277 & 0.6066 & 0.6303 & 0.4778 \\
Credit & Feature Only & KNN  & \textbf{0.9198} & 0.6725 & 0.5139 & 0.5776 & 0.4827 \\

Credit & Moderate Reg & XGB  & \textbf{0.8565} & 0.6139 & 0.5718 & 0.5756 & 0.4213 \\
Credit & Moderate Reg & TREE & \textbf{0.7943} & 0.5083 & 0.5277 & 0.4921 & 0.5023 \\
Credit & Moderate Reg & LR   & \textbf{0.8856} & 0.6931 & 0.5221 & 0.5817 & 0.4649 \\
Credit & Moderate Reg & NB   & \textbf{0.8685} & 0.6795 & 0.6019 & 0.6308 & 0.5015 \\
Credit & Moderate Reg & KNN  & \textbf{0.9000} & 0.5825 & 0.5442 & 0.5331 & 0.5035 \\

Credit  &  Drift & XGB  & \textbf{0.8870} & 0.6929 & 0.5576 & 0.6000 & 0.4662 \\
Credit &  Drift & TREE & \textbf{0.8249} & 0.5234 & 0.5402 & 0.5336 & 0.5084 \\
Credit  &  Drift & LR   & \textbf{0.8987} & 0.7501 & 0.5484 & 0.5892 & 0.4806 \\
Credit &  Drift & NB   & \textbf{0.9179} & 0.7636 & 0.5634 & 0.6564 & 0.4477 \\
Credit &  Drift & KNN  & \textbf{0.9276} & 0.6555 & 0.4925 & 0.5681 & 0.4755 \\

Credit & Strong Reg & XGB  & \textbf{0.8283} & 0.5844 & 0.6108 & 0.5584 & 0.4749 \\
Credit & Strong Reg & TREE & \textbf{0.7771} & 0.5223 & 0.5861 & 0.5227 & 0.4829 \\
Credit & Strong Reg & LR   & \textbf{0.8626} & 0.6139 & 0.5484 & 0.5352 & 0.5245 \\
Credit & Strong Reg & NB   & \textbf{0.8187} & 0.6904 & 0.5739 & 0.5779 & 0.4732 \\
Credit & Strong Reg & KNN  & \textbf{0.7864} & 0.5176 & 0.5387 & 0.5037 & 0.4965 \\

\bottomrule
\end{tabular}
\end{adjustbox}
\caption{Per-scenario APFD results for Hospital and Credit Card datasets.}
\label{tab:hospital_credit}
\end{table}

\begin{table}[!t]]
\centering
\scriptsize
\setlength{\tabcolsep}{2.3pt}
\begin{adjustbox}{max width=\textheight}
\begin{tabular}{lllccccc}
\toprule
Dataset & Scenario & Classifier & RiskBlend & DeepGini & DATIS & MLPrior & Random \\
\midrule

Bank & Feature Only & XGB  & \textbf{0.9899} & 0.9696 & 0.9152 & 0.9412 & 0.4929 \\
Bank & Feature Only & TREE & \textbf{0.9490} & 0.7722 & 0.8873 & 0.7196 & 0.4951 \\
Bank & Feature Only & LR   & \textbf{0.9687} & 0.9181 & 0.8553 & 0.9202 & 0.5072 \\
Bank & Feature Only & NB   & \textbf{0.9475} & 0.8786 & 0.8833 & 0.9158 & 0.5006 \\
Bank & Feature Only & KNN  & \textbf{0.9888} & 0.9295 & 0.9175 & 0.8690 & 0.4961 \\

Bank & Moderate Reg & XGB  & \textbf{0.9835} & 0.9326 & 0.8480 & 0.9149 & 0.5032 \\
Bank & Moderate Reg & TREE & \textbf{0.9405} & 0.7540 & 0.8621 & 0.7117 & 0.5039 \\
Bank & Moderate Reg & LR   & \textbf{0.9298} & 0.8549 & 0.8327 & 0.8857 & 0.5001 \\
Bank & Moderate Reg & NB   & \textbf{0.9421} & 0.7456 & 0.8384 & 0.8123 & 0.4985 \\
Bank & Moderate Reg & KNN  & \textbf{0.9653} & 0.7177 & 0.8546 & 0.7419 & 0.5028 \\

Bank &  Drift & XGB  & \textbf{0.9805} & 0.8859 & 0.8309 & 0.8809 & 0.4992 \\
Bank &  Drift & TREE & \textbf{0.9365} & 0.7611 & 0.8586 & 0.7224 & 0.5028 \\
Bank &  Drift & LR   & \textbf{0.9353} & 0.8018 & 0.8149 & 0.8832 & 0.5029 \\
Bank &  Drift & NB   & \textbf{0.9398} & 0.8577 & 0.8166 & 0.8454 & 0.4992 \\
Bank &  Drift & KNN  & \textbf{0.9637} & 0.6980 & 0.8558 & 0.7441 & 0.5021 \\

Bank & Strong Reg & XGB  & \textbf{0.9837} & 0.9286 & 0.8605 & 0.9149 & 0.5079 \\
Bank & Strong Reg & TREE & \textbf{0.9387} & 0.7654 & 0.8600 & 0.7154 & 0.5004 \\
Bank & Strong Reg & LR   & \textbf{0.9239} & 0.7964 & 0.8166 & 0.8604 & 0.5033 \\
Bank & Strong Reg & NB   & \textbf{0.9443} & 0.6937 & 0.8111 & 0.7973 & 0.4969 \\
Bank & Strong Reg & KNN  & \textbf{0.9628} & 0.7569 & 0.8518 & 0.7213 & 0.4991 \\

\midrule

Adult & Feature Only & XGB  & \textbf{0.9916} & 0.9616 & 0.8568 & 0.9062 & 0.5240 \\
Adult & Feature Only & TREE & \textbf{0.9211} & 0.7545 & 0.8250 & 0.6974 & 0.4995 \\
Adult & Feature Only & LR   & \textbf{0.8764} & 0.6904 & 0.8124 & 0.8502 & 0.5020 \\
Adult & Feature Only & NB   & \textbf{0.9955} & 0.9902 & 0.8922 & 0.8116 & 0.5036 \\
Adult & Feature Only & KNN  & \textbf{0.9675} & 0.8315 & 0.8597 & 0.7999 & 0.5079 \\

Adult & Moderate Reg & XGB  & \textbf{0.9901} & 0.9384 & 0.8448 & 0.8938 & 0.5091 \\
Adult & Moderate Reg & TREE & \textbf{0.9198} & 0.7532 & 0.8266 & 0.7038 & 0.5057 \\
Adult & Moderate Reg & LR   & \textbf{0.8460} & 0.6815 & 0.8042 & 0.8269 & 0.5004 \\
Adult & Moderate Reg & NB   & \textbf{0.9960} & 0.7740 & 0.9608 & 0.8600 & 0.5008 \\
Adult & Moderate Reg & KNN  & \textbf{0.9600} & 0.8243 & 0.8745 & 0.7879 & 0.5010 \\

Adult &  Drift & XGB  & \textbf{0.9897} & 0.9681 & 0.8502 & 0.9166 & 0.5096 \\
Adult &  Drift & TREE & \textbf{0.9253} & 0.7765 & 0.8122 & 0.7359 & 0.4992 \\
Adult &  Drift & LR   & \textbf{0.9184} & 0.7213 & 0.7797 & 0.7877 & 0.4999 \\
Adult &  Drift & NB   & \textbf{0.9991} & 0.9959 & 0.8933 & 0.8503 & 0.4839 \\
Adult &  Drift & KNN  & \textbf{0.9709} & 0.8450 & 0.8545 & 0.8308 & 0.5055 \\

Adult & Strong Reg & XGB  & \textbf{0.9767} & 0.8794 & 0.7853 & 0.8417 & 0.5052 \\
Adult & Strong Reg & TREE & \textbf{0.9144} & 0.7340 & 0.8086 & 0.6814 & 0.5040 \\
Adult & Strong Reg & LR   & \textbf{0.8427} & 0.7095 & 0.7847 & 0.8014 & 0.4980 \\
Adult & Strong Reg & NB   & \textbf{0.9668} & 0.6924 & 0.8124 & 0.7549 & 0.5046 \\
Adult & Strong Reg & KNN  & \textbf{0.9350} & 0.7398 & 0.8387 & 0.7489 & 0.5045 \\

\bottomrule
\end{tabular}
\end{adjustbox}
\caption{Per-scenario APFD results for Bank and Adult datasets.}
\label{tab:bank_adult}
\end{table}

\subsection{RQ2: Signal Contribution and Informative Signals}

To answer RQ2, we analyze the validation-learned signal weights produced
by RiskBlend's APFD-squared weighting mechanism across all four datasets,
scenarios, and classifiers. For each experimental configuration, the weight
assigned to each signal reflects its estimated contribution to
regression-fault detection in the held-out validation partition. We report the mean weights averaged on 15 seeds.
Table~\ref{tab:signal_weights_dataset} shows the mean weight per-dataset.
The boundary shift dominates in three of four datasets (Adult: 0.349, Bank:
0.324, Credit: 0.375). The hospital is the exception, where history risk
receives the highest weight (0.366), closely followed by boundary shift
(0.309). This reflects the structured clinical feature space of the hospital, where
the historical misclassification patterns of $V_k$ in training data (captured
by history risk) provide a particularly strong signal for identifying
regression-prone test inputs. Across all four datasets, neighborhood change
is consistently the weakest signal, ranging from 0.099 (Hospital) to 0.142
(Adult).

\begin{table}[h!]
\centering
\caption{Mean validation-learned weights per signal and dataset.
Bold indicates the dominant signal per dataset.}
\setlength{\tabcolsep}{4pt}
\begin{tabular}{lcccc}
\toprule
\textbf{Dataset} & \textbf{History} & \textbf{Boundary} &
\textbf{Pred.\ Shift} & \textbf{Neigh.} \\
\midrule
Adult    & 0.214 & \textbf{0.349} & 0.295 & 0.142 \\
Bank     & 0.230 & \textbf{0.324} & 0.306 & 0.141 \\
Credit   & 0.209 & \textbf{0.375} & 0.280 & 0.136 \\
Hospital & \textbf{0.366} & 0.309 & 0.225 & 0.099 \\
\bottomrule
\end{tabular}
\label{tab:signal_weights_dataset}
\end{table}

\begin{table}[t]
\centering
\caption{Leave-one-signal-out (LOSO) ablation study. Each value reports the mean APFD drop ($\Delta$APFD) after removing one RiskBlend signal. Larger positive values indicate greater contribution to regression-fault detection. Results are averaged over all scenarios, classifiers, and seeds.}
\label{tab:loso_ablation}
\begin{tabular}{lcccc}
\toprule
\textbf{Removed Signal} &
\textbf{Adult} &
\textbf{Bank} &
\textbf{Hospital} &
\textbf{Credit} \\
\midrule
Boundary Shift      & \textbf{0.0469} & \textbf{0.0328} & \textbf{0.0686} & \textbf{0.0749} \\
History Risk        & 0.0042 & 0.0021 & 0.0169 & 0.0232 \\
Neighborhood Change & $-$0.0003 & $-$0.0006 & $-$0.0012 & $-$0.0030 \\
Prediction Shift    & $-$0.0042 & $-$0.0025 & $-$0.0125 & $-$0.0233 \\
\bottomrule
\end{tabular}
\end{table}

Table~\ref{tab:loso_ablation} complements the learned-weight analysis by
quantifying the actual contribution of each signal to the prioritization
performance through a leave-one-signal-out (LOSO) ablation study. Unlike the
validation-learned weights, which indicate how strongly the optimizer relies
on each signal during weight learning, the LOSO analysis measures the
performance degradation after removing an individual signal while
re-normalizing the remaining weights. Across all four datasets, the removal of the
\emph{Boundary Shift} signal results in the largest reduction in APFD, with
mean drops ranging from 0.0328 (Bank) to 0.0749 (Credit Card), confirming
that changes in the decision boundary between $V_k$ and $V_{k+1}$ provide
the strongest evidence to identify regression-prone test inputs.
Removing \emph{History Risk} produces a smaller but consistently positive APFD
reduction, particularly for the Hospital (0.0169) and Credit Card (0.0232)
datasets, indicating that historical fault patterns provide complementary
information beyond boundary changes. In contrast, removing
\emph{Neighborhood Change} has virtually no effect on APFD across all
datasets, while removing \emph{Prediction Shift} slightly improves APFD on
average. These results suggest that the information captured by the
prediction shift is largely redundant with the Boundary Shift signal, whereas
Neighborhood Change contributes only marginal additional information under the
considered regression-testing scenarios.

\vspace{4pt}
\noindent\textbf{Answer to RQ2:} The boundary shift is the most influential component of \textsc{RiskBlend}, receiving the highest average learned weight ($\bar{w}=0.349$) and producing the largest APFD reduction in the leave-one-signal-out ablation across all datasets. The prediction shift is consistently assigned the second-highest weight ($\bar{w}=0.277$), while the history risk provides a stable complementary contribution ($\bar{w}=0.255$), particularly for the hospital data set. The neighborhood change receives the lowest average weight ($\bar{w}=0.130$) and has the least impact on the APFD, indicating that its information is largely redundant with the other signals. The learned weights remain stable across the regression scenarios ($\sigma<0.05$), demonstrating that the validation-based weighting strategy is consistently generalized across the datasets and classifier families.

\subsection{Statistical Significance}

To assess whether RiskBlend's APFD improvements are statistically
significant, we applied the Wilcoxon signed-rank test, pairing
RiskBlend against each baseline by (scenario, seed, classifier) per
dataset, yielding $N = 300$ paired observations per comparison
(4~scenarios $\times$ 15~seed $\times$ 5~classifiers). The $p$-values
are reported with the Bonferroni correction for four simultaneous
comparisons. The effect size is measured by Cliff's $\delta$, with
thresholds $|\delta| < 0.147$ negligible, $< 0.330$ small,
$< 0.474$ medium, and $\geq 0.474$ large~\cite{romano2006appropriate}.

All 16 comparisons in four datasets are statistically significant
at $p < 0.001$ after Bonferroni correction, with large effect sizes
in 15 of 16 cases. Against Random, RiskBlend achieves near-complete
stochastic dominance across all datasets ($\delta \geq 0.993$,
$\Delta\mathrm{APFD} \geq +0.380$). Against DATIS-Tabular, effect
sizes are also uniformly large ($\delta \geq 0.973$). Against
MLPrior, the effect sizes are large in all datasets ($\delta \geq 0.630$,
$\Delta\mathrm{APFD} \geq +0.130$). Against DeepGini, the strongest
baseline effect sizes are large in Hospital ($\delta = 0.567$),
Credit Card ($\delta = 0.898$) and Bank ($\delta = 0.887$). The
single exception is RiskBlend vs.\ DeepGini in Adult
($\delta = 0.084$, negligible, $\Delta\mathrm{APFD} = +0.077$),
where DeepGini remains competitive in linear classifiers with sparse
categorical features, reducing the practical margin despite
statistical significance. In Bank, RiskBlend exceeded every baseline
in all 300 paired observations, yielding the maximum possible
Wilcoxon statistic ($W = 45{,}150$) for all four comparisons.
The replication package is publicly available on Figshare.\footnote{\url{https://doi.org/10.6084/m9.figshare.32964359}}.

\section{Discussion}

The results confirm that the combination of multiple risk signals produces more
robust prioritization than any single-signal strategy under diverse
regression-update conditions. RiskBlend achieves the highest APFD in all
80 classifier scenario cells, with the largest gains on tree-based and
distance-based classifiers, where confidence-based methods lose
discriminative power after version transitions. Even on well-calibrated
classifiers where confidence is strongest, RiskBlend remains competitive,
demonstrating that multi-signal blending does not sacrifice performance in
favorable conditions while substantially improving it in adverse ones.
Confidence-based methods assume that model uncertainty reliably reflects
fault risk after a version update. Our results show this weakens for
Decision Tree classifiers across all datasets, where DeepGini APFD falls
to 0.547--0.623 while RiskBlend maintains 0.816--0.858. The weight
analysis (RQ2) explains this: The boundary shift and the prediction shift jointly
receive $\approx$62\% of the ensemble weight, compensating for the reduced
informativeness of the single-version confidence after a version transition.
MLPrior performs well on larger datasets, but degrades on smaller ones
(Credit Card), exposing a data-volume sensitivity that RiskBlend's
validation-learned weighting does not share.
No single signal consistently dominates across all configurations; rather,
signal importance adapts to the dataset and the classifier via validation-learned
weighting without classifier-specific tuning. Limitations include sensitivity
to the held-out validation partition for weight learning and reduced
effectiveness when regression faults are very sparse. Future work includes
extension to multi-class and regression settings, and online weight
adaptation across model version sequences.

Regression fault densities varied substantially across datasets, classifiers, and regression-update scenarios, with average fault rates ranging from approximately 3.6\% (Bank) to 14.0\% (Credit Card). This variability reflects differences in classifier behavior and update severity, resulting in both easy and challenging prioritization settings. Consequently, APFD values, particularly near-ceiling scores (e.g., $\geq 0.98$), should be interpreted in the context of the underlying regression fault density.

Table~\ref{tab:runtime} reports the average wall-clock execution time of
each prioritization method, averaged over all classifiers, random seeds,
and regression-update scenarios. The reported time measures only the
prioritization stage before oracle verification, excluding model training,
dataset preparation, and manual labeling. Random and DeepGini incur
negligible overhead (less than 0.001~s). DATIS requires between 0.62~s and
5.30~s across datasets. RiskBlend completes in 0.32--8.95~s, and MLPrior in
0.20--17.63~s. Relative to MLPrior, RiskBlend is substantially faster on
Bank (4.94~s vs.\ 17.63~s) but slower on Adult (8.95~s vs.\ 1.16~s),
Hospital (2.39~s vs.\ 0.76~s), and Credit Card (0.32~s vs.\ 0.20~s). These
results indicate that RiskBlend's computational cost is dataset-dependent,
and while it avoids MLPrior's worst-case overhead on Bank, it does not
consistently offer lower overhead than the mutation-based baseline.
\begin{table}[h]
\centering
\caption{Average execution time (seconds) per prioritization method across datasets
         (mean over all classifiers, seeds, scenarios}
\label{tab:runtime}
\begin{tabular}{lcccc}
\toprule
\textbf{Method}
  & \textbf{Adult}
  & \textbf{Bank}
  & \textbf{Hospital}
  & \textbf{Credit} \\
\midrule
Random        & 0.000199 & 0.000222 & 0.000164 & 0.000128 \\
DeepGini      & 0.000102 & 0.000120 & 0.000098 & 0.000038 \\
MLPrior   & 1.1590   & 17.6336  & 0.7584   & 0.2043   \\
DATIS & 4.0762   &  5.3016  & 1.5293   & 0.6218   \\
RiskBlend     & 8.9482   &  4.9425  & 2.3929   & 0.3152   \\
\bottomrule
\end{tabular}
\end{table}

\section{Threats to Validity}

\textbf{Internal validity:} Signal weights are learned from a held-out
20\% validation partition of $V_{k+1}$'s training data. If this partition
is small or unrepresentative of the regression-fault distribution, learned
weights may not generalize to the test set. We mitigate this by averaging
results over 15 random seeds, which stabilizes weight estimates across
partitioning choices.

\textbf{External validity:} We evaluate RiskBlend in four benchmark
datasets that span the financial, socioeconomic, and healthcare domains in
five classifier families and four regression-update scenarios. Although this
covers a broad range of conditions, results may not generalize to
unstructured data (images, text), deep learning models, or production
pipelines with continuous data drift. regression-update scenarios were
constructed to reflect realistic maintenance patterns; however, real-world
update distributions may differ in ways not captured by our four scenarios.

\section{Conclusion}

This work presented \textit{RiskBlend}, a classifier-agnostic framework
for prioritization of test input in ML regression testing. RiskBlend fuses
four generic risk signals, historical failure patterns, prediction shift,
decision-boundary shift, and neighborhood change, into a single score
via APFD-squared weighting learned from validation, requiring only no
classifier-specific tuning. Evaluated across four datasets, five
classifiers, and four regression-update scenarios (1{,}200 configurations
per dataset), RiskBlend achieves the highest APFD in all 80 classifier--scenario cells,
with statistically significant improvements over all baselines. The Signal-weight analysis identifies the boundary shift as the most
informative signal ($\bar{w} = 0.339$), with stable weights between 
scenarios and classifiers, confirming that the behavioral signals in cross-version
 are essential complements to the confidence of the single-model for the prioritization of the robust
 regression test ML.

\bibliographystyle{plain}
\bibliography{ref}

@article{dang2024ml,
  author    = {Dang, Xuan and Li, Yinghua and Papadakis, Mike and Klein, Jacques and Bissyand{\'e}, Tegawend{\'e} F. and Le Traon, Yves},
  title     = {Test Input Prioritization for Machine Learning Classifiers},
  journal   = {IEEE Transactions on Software Engineering},
  volume    = {50},
  number    = {3},
  pages     = {413--442},
  year      = {2024},
  doi       = {10.1109/TSE.2024.3350019}
}

@article{dang2024gnn,
  author    = {Dang, Xuan and Li, Yinghua and Papadakis, Mike and Klein, Jacques and Bissyand{\'e}, Tegawend{\'e} F. and Le Traon, Yves},
  title     = {Test Input Prioritization for Graph Neural Networks},
  journal   = {IEEE Transactions on Software Engineering},
  volume    = {50},
  number    = {6},
  year      = {2024},
  doi       = {10.1109/TSE.2024.3387002}
}

@article{graphrank2025,
  author    = {Yang, Lichuang and He, Daojing and Li, Yu and others},
  title     = {Toward Efficient Testing of Graph Neural Networks via Test Input Prioritization},
  journal   = {Automated Software Engineering},
  year      = {2025},
  doi       = {10.1007/s10515-025-00554-0}
}

@inproceedings{kim2019sadl,
  author    = {Kim, Jinhan and Feldt, Robert and Yoo, Shin},
  title     = {Guiding Deep Learning System Testing Using Surprise Adequacy},
  booktitle = {Proceedings of the 41st International Conference on Software Engineering (ICSE)},
  pages     = {1039--1049},
  year      = {2019},
  publisher = {IEEE Press},
  doi       = {10.1109/ICSE.2019.00108}
}

@inproceedings{wang2021prima,
  author    = {Wang, Zan and You, Hanmo and Chen, Junjie and Zhang, Yingyi and Dong, Xuyuan and Zhang, Wenbin},
  title     = {Prioritizing Test Inputs for Deep Neural Networks via Mutation Analysis},
  booktitle = {Proceedings of the 43rd IEEE/ACM International Conference on Software Engineering (ICSE)},
  pages     = {397--409},
  year      = {2021},
  publisher = {IEEE},
  doi       = {10.1109/ICSE43902.2021.00046}
}

@inproceedings{feng2020deepgini,
  author    = {Feng, Yang and Shi, Qingkai and Gao, Xinyu and Wan, Jun and Fang, Chunrong and Chen, Zhenyu},
  title     = {{DeepGini}: Prioritizing Massive Tests to Enhance the Robustness of Deep Neural Networks},
  booktitle = {Proceedings of the 29th ACM SIGSOFT International Symposium on Software Testing and Analysis (ISSTA)},
  pages     = {177--188},
  year      = {2020},
  publisher = {ACM},
  doi       = {10.1145/3395363.3397357}
}

@inproceedings{li2024datis,
  author    = {Li, Zhong and Xu, Zhengfeng and Ji, Ruihua and Pan, Minxue and Zhang, Tian and Wang, Linzhang and Li, Xuandong},
  title     = {Distance-Aware Test Input Selection for Deep Neural Networks},
  booktitle = {Proceedings of the 33rd ACM SIGSOFT International Symposium on Software Testing and Analysis (ISSTA)},
  year      = {2024},
  publisher = {ACM},
  doi       = {10.1145/3650212.3652125}
}

@inproceedings{hao2023mots,
  author    = {Hao, Yao and Huang, Zhiqiu and Guo, Hongjing and Shen, Guohua},
  title     = {Test Input Selection for Deep Neural Network Enhancement Based on Multiple-Objective Optimization},
  booktitle = {Proceedings of the IEEE International Conference on Software Analysis, Evolution and Reengineering (SANER)},
  pages     = {534--545},
  year      = {2023},
  publisher = {IEEE},
  doi       = {10.1109/SANER56733.2023.00060}
}

@article{wu2023ssoa,
  author    = {Wu, Pengcheng and others},
  title     = {Stratified Random Sampling for Neural Network Test Input Selection},
  journal   = {Information and Software Technology},
  volume    = {163},
  year      = {2023},
  doi       = {10.1016/j.infsof.2023.107290}
}

@Article{electronics13173380,
AUTHOR = {Srinivasan, Madhusudan and Kanewala, Upulee},
TITLE = {Improving Early Fault Detection in Machine Learning Systems Using Data Diversity-Driven Metamorphic Relation Prioritization},
JOURNAL = {Electronics},
VOLUME = {13},
YEAR = {2024},
NUMBER = {17},
ARTICLE-NUMBER = {3380},
URL = {https://www.mdpi.com/2079-9292/13/17/3380},
ISSN = {2079-9292},
DOI = {10.3390/electronics13173380}
}

@inproceedings{byun2019input1,
  title={Input prioritization for testing neural networks},
  author={Byun, Taejoon and Sharma, Vaibhav and Vijayakumar, Abhishek and Rayadurgam, Sanjai and Cofer, Darren},
  booktitle={2019 IEEE International Conference On Artificial Intelligence Testing (AITest)},
  pages={63--70},
  year={2019},
  organization={IEEE}
}

@inproceedings{pan2022test,
  title={Test case prioritization for deep neural networks},
  author={Pan, Zhonghao and Zhou, Shan and Wang, Jianmin and Wang, Jinbo and Jia, Jiao and Feng, Yang},
  booktitle={2022 9th International conference on dependable systems and their applications (DSA)},
  pages={624--628},
  year={2022},
  organization={IEEE}
}

@inproceedings{romano2006appropriate,
  title={Appropriate statistics for ordinal level data: Should we really be using t-test and Cohen’sd for evaluating group differences on the NSSE and other surveys},
  author={Romano, Jeanine and Kromrey, Jeffrey D and Coraggio, Jesse and Skowronek, Jeff},
  booktitle={annual meeting of the Florida Association of Institutional Research},
  volume={177},
  number={34},
  year={2006}
}
\end{document}